\pdfoutput=1

\documentclass[11pt]{article}

\usepackage[preprint]{acl}
\usepackage{paralist} 

\usepackage{enumitem} 
\usepackage{paralist} 
\usepackage{times}         
\usepackage{latexsym}      
\usepackage{microtype}     
\usepackage{subfigure}     
\usepackage{booktabs}      
\usepackage{hyperref}      
\usepackage{amsmath}       
\usepackage{amssymb}       
\usepackage{mathtools}     
\usepackage{amsthm}        
\usepackage{algpseudocode} 
\usepackage{multirow}      
\usepackage{bbm}           
\usepackage{colortbl}      
\usepackage{xspace}        
\usepackage{wrapfig}       
\usepackage{arydshln}      
\usepackage{makecell}      
\usepackage{tcolorbox}     
\tcbuselibrary{breakable}
\usepackage{listings}      
\usepackage{url}           
\usepackage{amsfonts}      
\usepackage{nicefrac}      
\usepackage{xcolor}        
\usepackage{pifont}        
\usepackage{CJKutf8}       
\usepackage{amsmath}
\DeclareMathOperator*{\argmax}{argmax}

\newcommand {\ourmethod}[1]{CEP}
\usepackage[T1]{fontenc}

\usepackage[utf8]{inputenc}

\newcommand{\model}{\textsc{C-Prune}}

\usepackage{inconsolata}

\usepackage{graphicx}

\title{Cluster-Driven Expert Pruning for Mixture-of-Experts Large Language Models}

  \author{
 \textbf{Hongcheng Guo}\textsuperscript{1}\textsuperscript{,}\thanks{Equal contribution.},
 \textbf{Juntao Yao}\textsuperscript{2}\textsuperscript{,}\footnotemark[1],
 \textbf{Boyang Wang}\textsuperscript{1},\\
 \textbf{Junjia Du}\textsuperscript{3}, 
 \textbf{Shaosheng Cao}\textsuperscript{4}\textsuperscript{,}\footnotemark[2],
 \textbf{Donglin Di}\textsuperscript{5},
 \textbf{Shun Zhang}\textsuperscript{1},
 \textbf{Zhoujun Li}\textsuperscript{1}\textsuperscript{,}\thanks{Corresponding author.}
 \\
 \\
 \textsuperscript{1}Beihang University,
 \textsuperscript{2}University of Washington,\\
 \textsuperscript{3}Nanyang Technological University,
 \textsuperscript{4}Xiaohongshu Inc.
 \textsuperscript{5}Tsinghua University
}

\begin{document}
\maketitle
\begin{abstract}
Mixture-of-Experts (MoE) architectures have emerged as a promising paradigm for scaling large language models (LLMs) with sparse activation of task-specific experts. Despite their computational efficiency during inference, the massive overall parameter footprint of MoE models (e.g., GPT-4) introduces critical challenges for practical deployment. Current pruning approaches often fail to address two inherent characteristics of MoE systems: 1).intra-layer expert homogeneity where experts within the same MoE layer exhibit functional redundancy, and 2). inter-layer similarity patterns where deeper layers tend to contain progressively more homogeneous experts. To tackle these issues, we propose \textbf{C}luster-driven Expert Pruning (\model{}), a novel two-stage framework for adaptive task-specific compression of MoE LLMs. \model{} operates through layer-wise expert clustering, which groups functionally similar experts within each MoE layer using parameter similarity metrics, followed by global cluster pruning, which eliminates redundant clusters across all layers through a unified importance scoring mechanism that accounts for cross-layer homogeneity. We validate \model{} through extensive experiments on multiple MoE models and benchmarks. The results demonstrate that \model{} effectively reduces model size while outperforming existing MoE pruning methods~\footnote{We provide code.
\url{https://github.com/Fighoture/MoE_unsupervised_pruning}
}.

\end{abstract}

\section{Introduction}

\begin{figure*}
    \centering
    \includegraphics[width=1\linewidth]{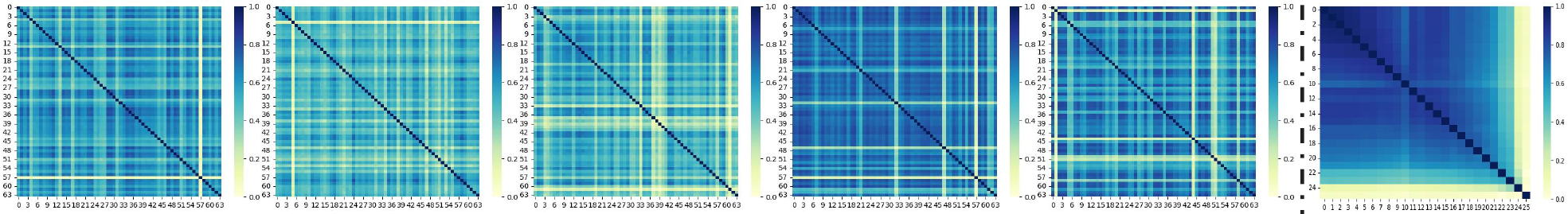}
    \caption{Visualization of expert cosine similarity in DeepSeek-V2-Lite based on math subject samples. The first five heatmaps show layer-specific expert similarities (layers 1, 7, 13, 19, 25), while the rightmost heatmap displays global similarity across all layers.}
    \label{fig:overview}
\end{figure*}

\begin{quote}
\large\textit{``The true art of model compression is not merely reducing parameters, but preserving functionality while achieving efficiency.''} \textbf{-- Inspired by Carl Jung}
\end{quote}

The Mixture-of-Experts (MoE) paradigm, first conceptualized in early modular networks~\cite{moe2024survey}, has evolved into a cornerstone for scaling large language models (LLMs) through sparse expert activation. Initial implementations in RNNs~\cite{shazeer2017outrageously} demonstrated its potential, while subsequent adaptations to Transformer architectures~\cite{lepikhin2020gshard, seer_moe, efficient_moe,owl} and decoder-only GPT variants~\cite{llama_moe, decoder_o, jiang2024mixtral} have established MoE as a mainstream approach for balancing performance and computational cost. However, the exponential growth of MoE model parameters (e.g., trillion-scale models) creates a critical deployment paradox: while inference activates only subsets of experts, the full parameter footprint remains prohibitive for real-world applications.

Existing compression efforts face two fundamental limitations. First, while expert pruning has shown promise in specialized domains like machine translation~\cite{moe_translate}—where language-specific experts can be selectively removed~\cite{diversify_moe}—these methods rely heavily on task-specific signals (e.g., gate activation statistics~\cite{seer_moe}) or require costly retraining pipelines~\cite{chen2022task}, making them impractical for general-purpose LLMs. Second, current approaches neglect the intrinsic structural properties of MoE models: I. Intra-layer homogeneity: Experts within the same layer frequently develop functional overlap due to training dynamics~\cite{lin2024moe}. II. Inter-layer similarity: Deeper layers exhibit progressively redundant expert patterns~\cite{liu2024moe}.
As evidenced by recent analyses~\cite{moe_RBench, OpenMoE}, this hierarchical redundancy renders conventional pruning strategies—which treat experts as independent units—both inefficient and performance-degrading, as shown in Figure~\ref{fig:overview}.




To address these challenges, Building on insights from modular network analysis~\cite{moe2024survey} and task-specific compression~\cite{LocMoE}, we propose Cluster-driven Expert Pruning (\model{}), \model{} leverages the inherent structure of MoE models through two key steps: (1) \textit{Layer-wise Clustering}, which groups functionally similar experts within Homogeneity-aware layers using parameter space analysis, extending beyond simple activation counting~\cite{diversify_moe}; and (2) \textit{Global Clustering Optimization}, which globally prunes redundant clusters across layers while preserving depth-specific functionality, overcoming the limitations of layer-isolated approaches in prior work~\cite{fedus2022switch}. By combining these strategies, \model{} effectively reduces redundancy while preserving the task-specific functionality essential for maintaining strong model performance.

We validate \model{} through extensive experiments on multiple MoE variants (e.g., DeepSeek-MoE) and benchmarks, demonstrating its effectiveness in achieving significant parameter reduction (25-35\%) without compromising performance. Our results highlight that \model{} outperforms existing pruning methods, particularly in low-compression regimes, and provides insights into the depth-dependent homogeneity trends of MoE models. The key contributions include:

\begin{itemize}
\item The first self-adaptive systematic framework addressing both intra-layer and inter-layer redundancy in MoE LLMs, validated through theoretical analysis and empirical studies.
\item A task-specific pruning methodology that outperforms task-agnostic approaches~\cite{moe_translate}, while maintaining generalizability.
\item Empirical evidence proves the effect of \model{} and challenges the assumption of layer-independent expert utility, revealing depth-dependent homogeneity trends.
\end{itemize}

\section{Related Work}
\subsection{Mixture-of-Experts Models}
First introduced in~\cite{moe2024survey, lin2024moe, liu2024moe}, a Mixture-of-Experts (MoE) model contains multiple separate networks, and each network processes a subset of the entire dataset. This separation can be viewed as a modular transformation of a multi-layer network. MoE structure is used for designing Recurrent Neural Networks (RNNs) in~\cite{shazeer2017outrageously} and further extended to encoder-decoder Transformer-based models~\cite{lepikhin2020gshard, seer_moe, efficient_moe}. With the recent development of decoder-only GPT family of models~\cite{llama_moe, decoder_o,decoder_t, decoder_u}, MoE models based on this structure gain popularity~\cite{jiang2024mixtral}. In this paper, we focus on post-training expert pruning/skipping methodologies for MoE LLMs.

\subsection{Expert Pruning for MoE Models}
Expert pruning within MoE models has garnered attention in the realm of Natural Language Processing~\cite{moe_RBench, OpenMoE, LocMoE, cao}, particularly in machine translation tasks~\cite{moe_translate}. In these contexts, the translation of specific languages often renders the expertise of other language specialists superfluous. The most activated experts are reserved in~\citet{diversify_moe} to prune a machine translation MoE model, and~\citet{seer_moe, efficient_moe} proposes expert pruning metrics based on gate statistics collected during decoding. Although these methods actively deal with expert pruning for MoE models, they are still limited to the machine translation domain with linguistic models. Researchers in~\cite{chen2022task} provide a dropping-while-training method that progressively drops the non-professional experts for target downstream tasks, and experiments are carried out on Switch Transformers models~\cite{fedus2022switch}. However, in the LLM era, it is usually difficult to afford such a training paradigm~\cite{survey_chatgpt, llm_survey, llms_survey}.


\section{Methodology}

\subsection{Task Definition}
The expert pruning task can be formulated as a multi-objective optimization problem:

\begin{equation}
\small
\begin{aligned}
\min_{\{\hat{\Theta}^l\}} & \underbrace{\mathbb{E}_{(x,y)\sim\mathcal{D}} \mathcal{L}(\hat{\mathcal{M}}(x;\hat{\mathcal{F}}),y)}_{\text{Task Loss}} \\
& + \lambda_1 \underbrace{\sum_{l=1}^L \text{Sim}(\Theta^l \setminus \hat{\Theta}^l)}_{\text{Similarity Constraint}} \\
& + \lambda_2 \underbrace{\sum_{l=1}^L \|\hat{W}^l\|_{2,1}}_{\text{Sparsity Penalty}}
\end{aligned}
\end{equation}

where $\scriptstyle \text{Sim}(S) = \frac{1}{|S|^2}\sum_{i,j\in S}\rho_{ij}$ measures intra-set similarity, and $\scriptstyle \|\cdot\|_{2,1}$ enforces column-wise sparsity in routing matrices.

\subsection{Progressive Pruning Framework}
Our method operates through two coordinated phases:

\paragraph{Phase 1: Layerwise Redundancy Reduction}
For each MoE layer $l$:

\begin{equation}
\small
\begin{split}
\mathcal{L}_l &= \underbrace{\mathbb{E}_x\left[\|F^l(x)-\hat{F}^l(x)\|_2\right]}_{\text{Function Preservation}} \\
&\quad + \gamma \underbrace{\sum_{i<j\in s^l} \rho_{ij}}_{\text{Redundancy Penalty}} \\
&\quad + \beta \underbrace{\text{KL}(p_{\text{orig}}^l(y|x) \| p_{\text{pruned}}^l(y|x))}_{\text{Distribution Alignment}}
\end{split}
\end{equation}

where $s^l$ denotes experts scheduled for pruning in layer $l$.

\paragraph{Phase 2: Global Consistency Preservation}
After layerwise pruning:
\begin{equation}
\small
\mathcal{L}_{\text{global}} = \sum_{l=1}^L \left( \underbrace{\mathbb{E}_x[\text{Cov}(\{\hat{f}_n^l(x)\})]}_{\text{Diversity Maintenance}} + \eta \underbrace{\|\hat{\mathcal{F}}\|_F^2}_{\substack{\text{Model}\\\text{Compactness}}} \right)
\end{equation}

\subsection{Similarity-Aware Pruning}
\paragraph{Expert Embedding} For expert $f_i$ in layer $l$, compute its characteristic embedding:
\begin{equation}
\small
\phi(f_i) = \mathbb{E}_{x\sim\mathcal{D}} \left[ \frac{1}{K}\sum_{k=1}^K f_i(x_k) \right] \in \mathbb{R}^d
\end{equation}

\paragraph{Adaptive Clustering} Define the merging criterion through spectral analysis:
\begin{equation}
\small
\mathcal{C}_k = \left\{ f_j \big| \|\phi(f_j) - \mu_k\|_2 < \tau^{(l)} \right\}
\end{equation}
where cluster threshold $\tau^{(l)}$ adapts to layer depth:
\begin{equation}
\small
\tau^{(l)} = \frac{1}{N}\sum_{i=1}^N \|\phi(f_i) - \bar{\phi}\|_2 + \delta \cdot \sigma^{(l)}
\end{equation}
with $\bar{\phi}$ being the centroid of all experts and $\sigma^{(l)}$ the embedding standard deviation.

\subsection{Dynamic Pruning Algorithm}
\begin{enumerate}
\item Compute expert affinity matrix:
\begin{equation}
\small
A_{ij} = \sigma\left(\alpha \cdot \frac{\phi(f_i)^\top \phi(f_j)}{\|\phi(f_i)\|\|\phi(f_j)\|}\right)
\end{equation}
where $\alpha$ controls similarity sensitivity.

\item Initialize clusters $\mathcal{C}_k = \{f_k\}, \forall k$

\item While $|\mathcal{C}| > N-r$:
\begin{align}
\small
(u^*, v^*) &= \argmax_{u,v} A_{uv} \\\scriptstyle
\mathcal{C}_{\text{new}} &= \mathcal{C}_u \cup \mathcal{C}_v \\ \scriptstyle
A_{\text{new}} &= \frac{|\mathcal{C}_u|A_u + |\mathcal{C}_v|A_v}{|\mathcal{C}_u| + |\mathcal{C}_v|}
\end{align}

\item Prune experts via:
\begin{equation}
\small
s^l = \left\{ f_j \big| \min_{c\in\mathcal{C}_{\text{keep}}} \|\phi(f_j) - \mu_c\|_2 > \zeta^{(l)} \right\}
\end{equation}
where $\zeta^{(l)}$ is the layer-specific pruning radius.
\end{enumerate} 

\subsection{Parameterized Expert Merging}
For each final cluster $\mathcal{C}_k$:
\begin{equation}
\small
\hat{\theta}_k = \sum_{f_i\in\mathcal{C}_k} \omega_i \theta_i, \quad \omega_i = \frac{\exp(\gamma \cdot A_{ik})}{\sum_{j\in\mathcal{C}_k} \exp(\gamma \cdot A_{jk})}
\end{equation}
with temperature $\gamma$ controlling fusion sharpness.

\subsection{Routing Policy Adaptation}
Update routing weights for merged experts:
\begin{equation}
\small
\hat{W}_k = \frac{1}{|\mathcal{C}_k|} \sum_{f_i\in\mathcal{C}_k} W_i + \epsilon \cdot \mathcal{N}(0,I)
\end{equation}
where $\epsilon$ controls exploration noise for routing diversity.

\section{Experiment}

\subsection{Experiment Setting}
\paragraph{Models and Infrastructure} We used DeepseekV2Lite (1 standard FFN + 26 MoE FFN layers) and Qwen1.5-MoE-A2.7B (24 MoE FFN layers) as our base models~\cite{deepseekv2, qwen_moe}. All experiments were conducted on a cluster of 32 NVIDIA A100 (80GB) GPUs. The hyperparameters are shown in Table~\ref{tab:hyperparams}.
\paragraph{Evaluation Protocol} Our evaluation covers three major benchmarks: MMLU~\cite{mmlu}, GSM8K~\cite{gsm8k}, and HumanEval~\cite{humaneval}, spanning computer science, mathematics, and business domains. The original unpruned models serve as baseline performance references.
\subsection{Main Experiments}
\noindent \textbf{Efficient Pruning with Performance Balance}  
With a 20\% pruning rate, C-Prune reduces the parameter count of the DeepSeek model from 15.7B to 13.0B, while the MMLU composite score decreases by only 1.4\%, significantly outperforming random pruning (64\% performance drop). For the Qwen model, parameters are compressed from 14.3B to 11.8B, retaining 88\% of the MMLU score, as shown in Table~\ref{tab:repobench}.

\noindent \textbf{Robustness Across Domain-Specific Tasks}  
On computer science tasks, the pruned DeepSeek model achieves a score of 51.50, far surpassing baseline methods (e.g., Group\&Merge: 33.50). For mathematical reasoning, C-Prune outperforms the original model (DeepSeek: 33.56 vs. 32.21). In HumanEval, scores reach 18.90 (DeepSeek) and 32.90 (Qwen), highlighting advantages in technical domains.

\noindent \textbf{Limitations of Baseline Methods}  
Random pruning nearly fails on GSM8K tasks. While Group\&Merge approaches C-Prune in Qwen's business tasks, its overall performance gap remains significant (average score: 30.45 vs. 38.75), reflecting insufficient global optimization in existing methods.

\noindent \textbf{Gains from Task-Specific Fine-Tuning}  
Task-specific optimization mitigates performance loss effectively. For example, the pruned Qwen model achieves 39.40 on GSM8K (vs. 53.58 for the base model), a 56\% improvement over non-fine-tuned methods (Group\&Merge: 25.38), demonstrating deployment flexibility.

\noindent \textbf{Cross-Architecture Generalization}  
C-Prune maintains superior performance across both DeepSeek and Qwen. HumanEval scores remain close to base models (Qwen: 32.90 vs. 49.40), validating generalization capabilities across heterogeneous MoE architectures.

\begin{table*}[ht]
\centering
\resizebox{0.9\textwidth}{!}{%
\begin{tabular}{cccccccccccc}
\toprule
\multirow{2}{*}{Method} & \multirow{2}{*}{Base Model} & \multirow{2}{*}{Parameters} & \multirow{2}{*}{Total Pruning Rate} & \multirow{2}{*}{\# of Routed Experts} & \multicolumn{4}{c}{MMLU} & \multirow{2}{*}{GSM8K} & \multirow{2}{*}{HumanEval} & \multirow{2}{*}{Average} \\ 
\cmidrule(lr){6-9}
 & & & & & Computer Science & Math & Business & Average & & & \\ 
\midrule
Base & DeepSeek-V2-Lite & 15.7B & 0 & 64 & 53.00 & 32.21 & 49.54 & 45.58 & 30.94 & 32.30 & 36.27 \\ 
Random & DeepSeek-V2-Lite & 13.0B & 0.2 & 52 & 19.00 & 12.32 & 17.53 & 16.28 & 0.057 & 0 & 5.446 \\ 
Seer Prune & DeepSeek-V2-Lite & 13.0B & 0.2 & 52 & 29.00 & 26.54 & 30.09 & 28.76 & 2.058 & 0 & 10.27 \\ 
Group\&Merge & DeepSeek-V2-Lite & 13.0B & 0.2 & 52 & 33.50 & 24.65 & 31.64 & 32.03 & 3.963 & 1.20 & 12.40 \\ 
\model{}(Ours) & DeepSeek-V2-Lite & 13.0B & 0.2 & 52 & \textbf{51.50} & \textbf{33.56} & \textbf{48.16} & \textbf{44.94} & \textbf{26.45} & \textbf{18.90} & \textbf{30.10} \\
\midrule
Base & Qwen1.5-MoE-A2.7B & 14.3B & 0 & 60 & 47.68 & 34.03 & 52.45 & 45.82 & 53.58 & 49.40 & 47.16 \\ 
Random & Qwen1.5-MoE-A2.7B & 11.8B & 0.2 & 48 & 14.50 & 13.81 & 11.04 & 13.12 & 10.44 & 12.90 & 12.15 \\ 
Seer Prune & Qwen1.5-MoE-A2.7B & 11.8B & 0.2 & 48 & 29.00 & 25.54 & 15.10 & 22.05 & 15.32 & 26.20 & 22.20 \\ 
Group\&Merge & Qwen1.5-MoE-A2.7B & 11.8B & 0.2 & 48 & 35.50 & 19.61 & \textbf{40.93} & 33.29 & 25.38 & 28.00 & 30.45 \\ 
\model{}(Ours) & Qwen1.5-MoE-A2.7B & 11.8B & 0.2 & 48 & \textbf{48.00} & \textbf{31.98} & 40.15 & \textbf{40.06} & \textbf{39.40} & \textbf{32.90} & \textbf{38.75} \\
\bottomrule
\end{tabular}%
}
\caption{Results of Model Evaluation on Benchmarks}
\label{tab:repobench}
\end{table*}

\section{Analysis}

\subsection{Layerwise vs. Global}
\begin{figure}
    \centering
    \includegraphics[width=.48\textwidth]{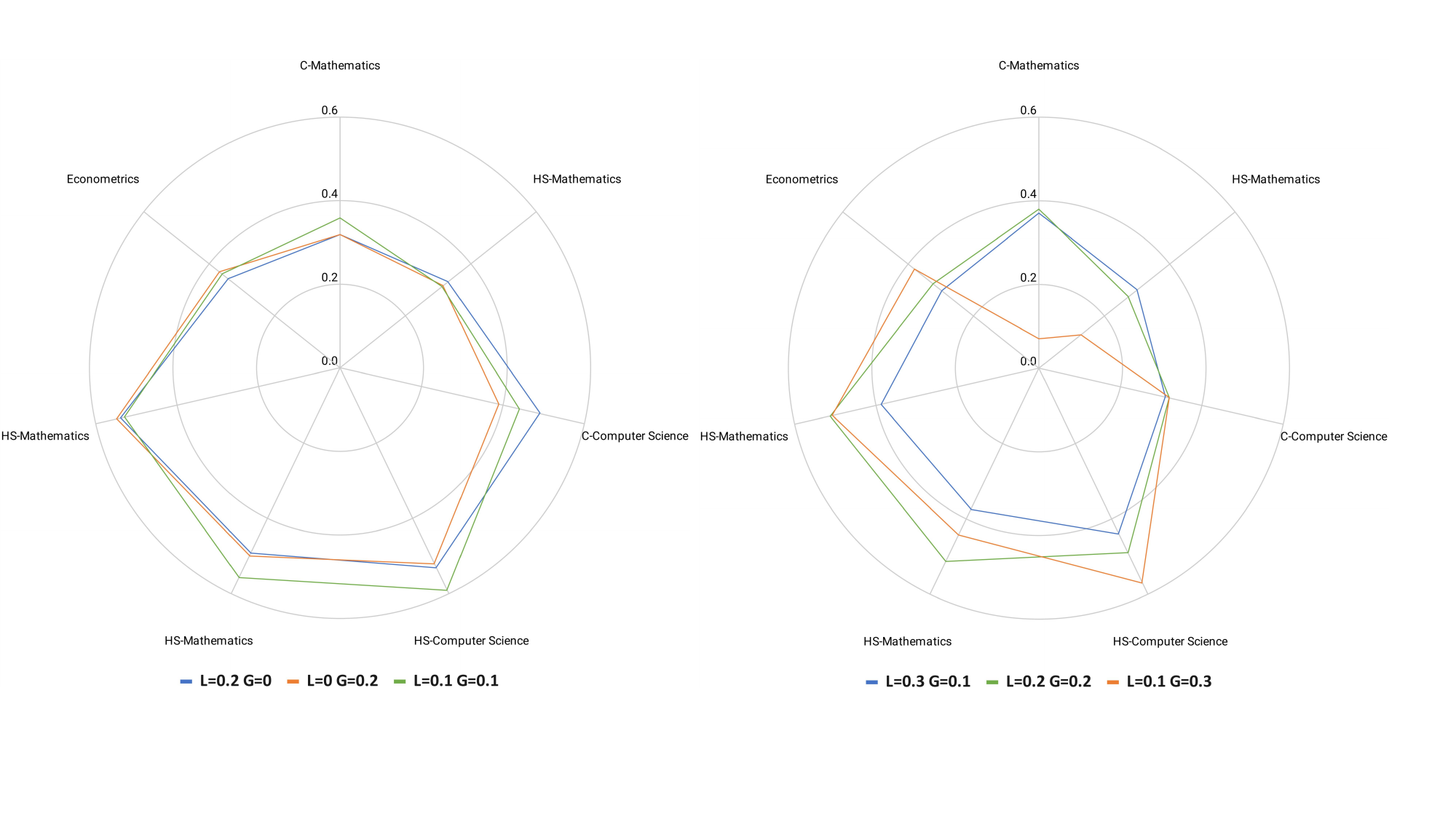}
    \caption{Performance comparisons across different academic subjects with varying Layer and Global pruning ratios.}
    \label{fig:radar_charts}
\end{figure}

\begin{figure}[t]
    \centering
    \begin{minipage}{0.22\textwidth}
        \centering
        \includegraphics[width=\linewidth]{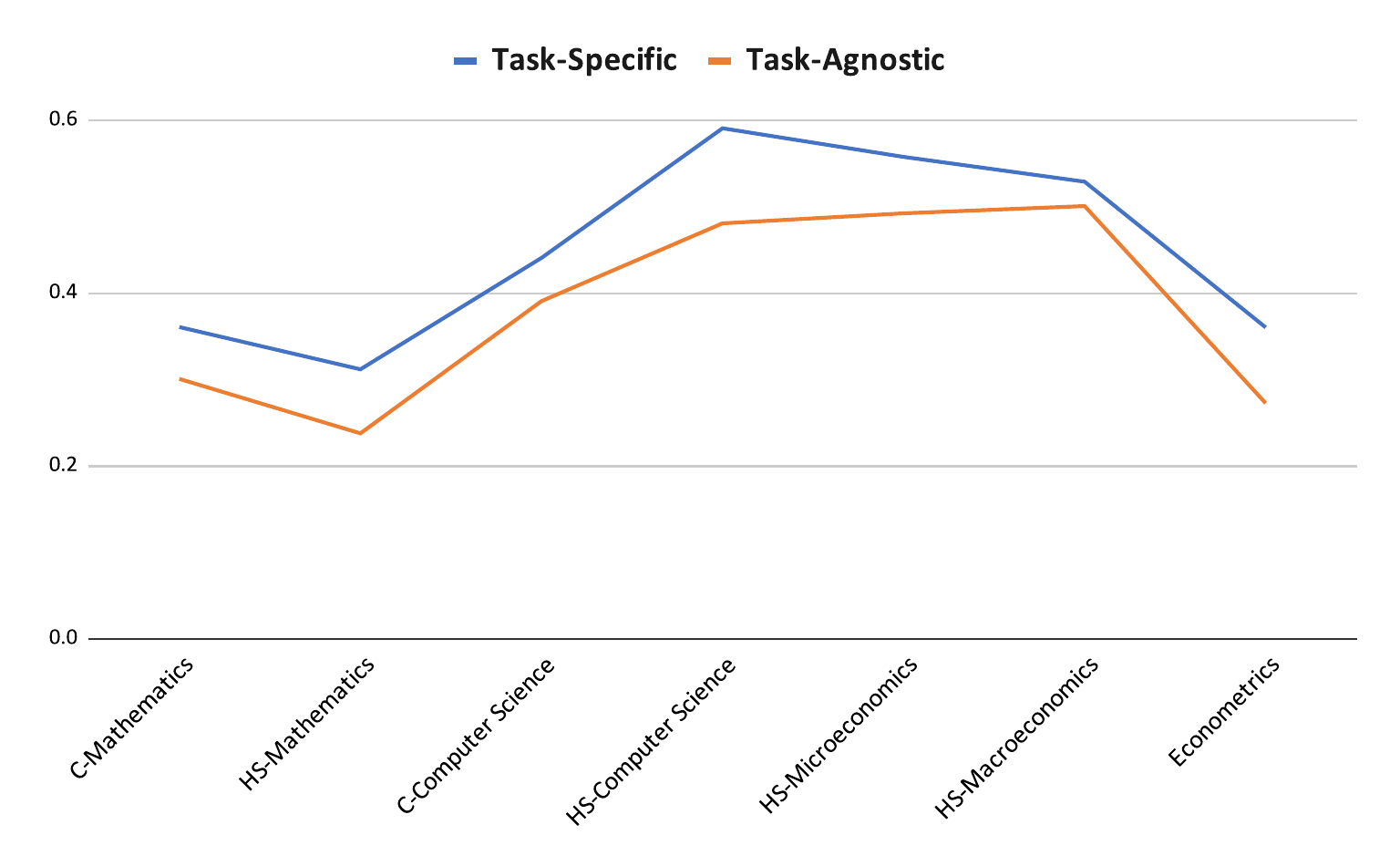}
        \caption{Performance comparison between Task‑Specific and Task‑Agnostic across different subject domains.}
        \label{fig:task_comparison}
    \end{minipage}
    \begin{minipage}{0.22\textwidth}
    \centering
    \includegraphics[width=\linewidth]{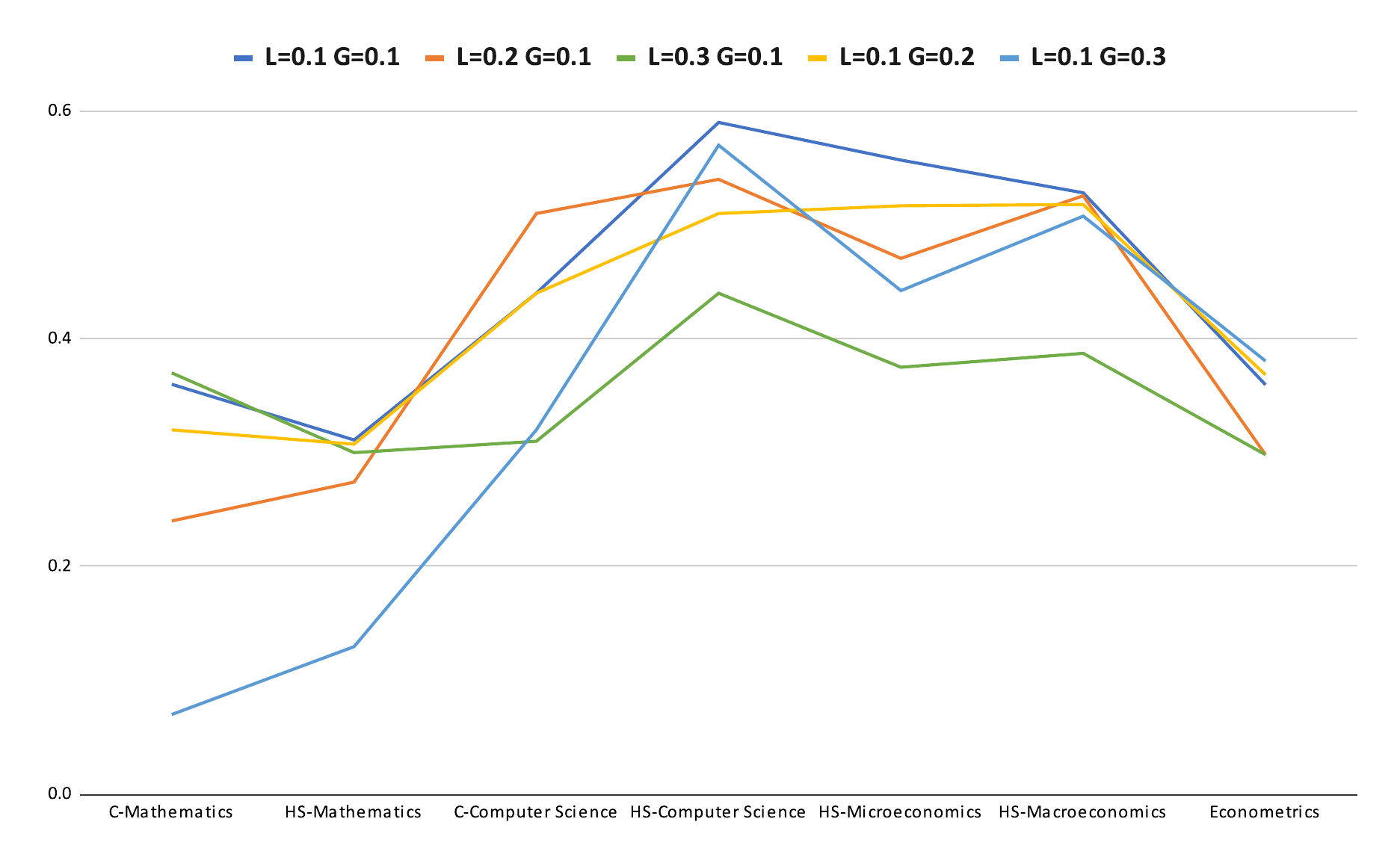}
    \caption{Performance comparison across different subject domains with varying Layer and Global pruning ratios.}
    \label{fig:pruning_ratios}
    \end{minipage}
    \vspace{-1.2em}
    \label{fig:combine_1}
\end{figure}
We conducted a systematic analysis of \textbf{\textit{Layer}} (L) and \textbf{\textit{Global}} (G) pruning effects across academic domains. The radar charts reveal a clear pattern where technical subjects show distinct responses to different pruning strategies. Specifically, Figure \ref{fig:radar_charts} (left) shows that when applying lower pruning ratios, subjects like mathematics and computer science maintain better performance under \textbf{\textit{Layerwise}} pruning, while Figure \ref{fig:radar_charts} (right) further validates this finding with higher pruning ratios, where economics exhibits more resilience to \textbf{\textit{Global}} pruning approaches. This differential response across domains, visualized through the radar patterns, suggests that knowledge organization within the model varies by subject matter, with technical knowledge being more layer-specific and general knowledge more distributed.

\subsection{Task-Agnostic vs. Task-Specific}
Figure \ref{fig:task_comparison} demonstrates the comparative effectiveness of task-specific versus task-agnostic pruning across academic domains. The task-specific approach consistently outperforms task-agnostic pruning, with the most pronounced advantage in computer science (\textbf{\textit{0.59}} vs \textbf{\textit{0.48}} at high school level). While mathematics shows smaller performance gaps between approaches, suggesting universal preservation of mathematical reasoning capabilities, computer science exhibits the highest absolute performance and largest benefit from specialized pruning. Economics maintains stable performance across both strategies, indicating its reliance on general language understanding. College-level subjects, particularly mathematics (\textbf{\textit{0.35}} task-specific, \textbf{\textit{0.30}} task-agnostic), show lower performance than their high school counterparts, highlighting the challenge of preserving advanced domain knowledge during pruning. These findings emphasize the importance of domain-aware pruning strategies, particularly for technically demanding subjects.

\subsection{Cross-Task Analysis}
\begin{figure}
    \centering
    \includegraphics[width=.4\textwidth]{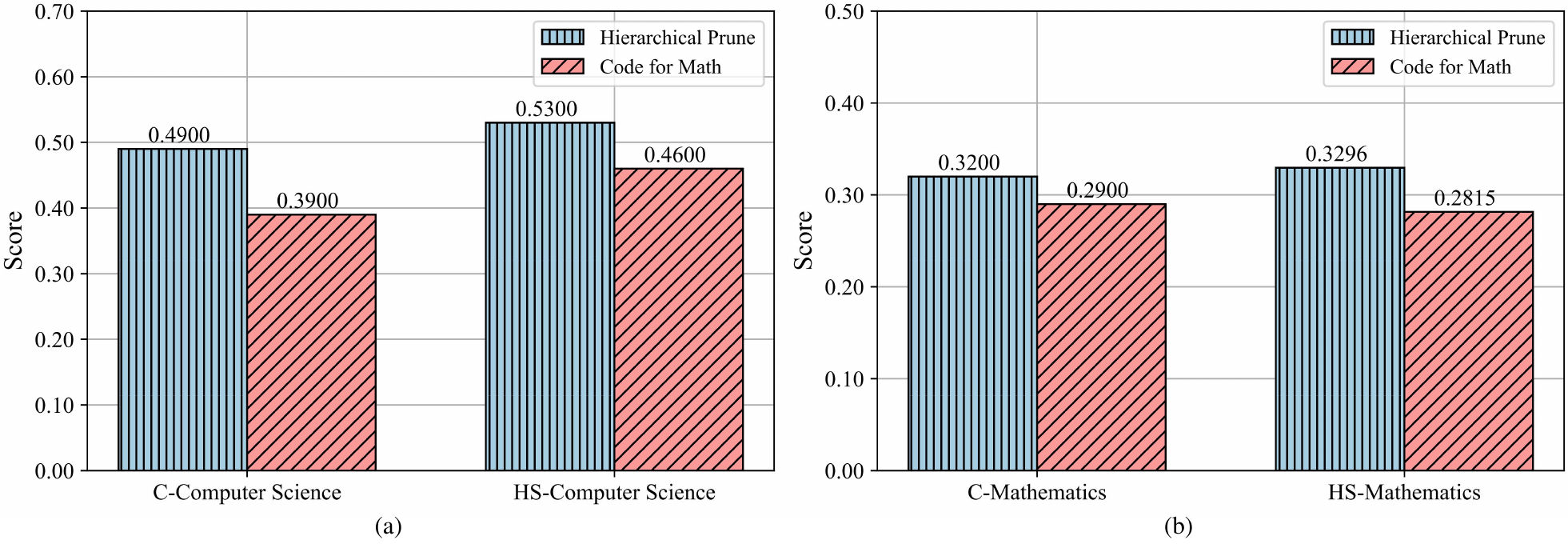}
    \caption{Performance comparison of Hierarchical Prune and Code for Math approaches across education levels.}
    \vspace{-1.2em}
    \label{fig:subject_comparison}
\end{figure}
Our investigation compared Hierarchical Prune with two task-specific methods - \textbf{\textit{Code for Math and Math for Code}} - to evaluate cross-domain transfer effectiveness. Using standardized scores [0,1], Figures \ref{fig:subject_comparison}
reveal that Hierarchical Prune maintained consistent performance across domains (computer science: college 0.70, high school 0.53; mathematics: college 0.50, high school 0.40). In contrast, task-specific methods showed significant degradation when transferred: \textbf{\textit{Code for Math}} performed poorly in mathematics (HS: 0.29), while \textbf{\textit{Math for Code}} struggled with computer science tasks (HS: 0.39), compared to their performance in native domains. These results demonstrate that domain adaptation requires careful consideration of both subject characteristics and educational complexity, as direct transfer of specialized methods leads to substantial performance decline.

\subsection{Pruning Ratios}
We systematically investigate the impact of pruning strategies on model performance across diverse academic domains. As shown in Figure \ref{fig:pruning_ratios}, we evaluate varying pruning ratios for both \textbf{\textit{Global}} and \textbf{\textit{Layerwise}} approaches to analyze the trade-off between model compression and performance retention. Through extensive experiments, we find that \textbf{\textit{economics-related tasks}} exhibit higher performance volatility under aggressive pruning parameters. In contrast, \textbf{\textit{computer science tasks}} demonstrate robust performance under moderate pruning configurations with Layer ratio 0.2 and Global ratio 0.1. The observed performance differential between educational levels within identical domains suggests that both knowledge complexity and domain characteristics significantly influence pruning efficacy. Our empirical analysis identifies optimal pruning configurations with \textbf{\textit{Global ratios}} between 0.1-0.2 and \textbf{\textit{Layerwise}} ratio approximately 0.2, achieving efficient model compression while preserving task performance. These findings provide insights for potential integration with complementary optimization techniques such as quantization and knowledge distillation to further enhance deployment efficiency.

\subsection{Number of Experts}
\begin{table}[h]
\centering
\resizebox{0.42\textwidth}{!}{%
\begin{tabular}{lccccc}
\toprule
Experts (Layerwise / Global) & 12 / 6 & 12 / 12 & 6 / 12 & 18 / 12 & 12 / 18 \\
\midrule
C-Mathematics       & \textbf{0.360}  & 0.290   & 0.310  & 0.310   & 0.350  \\
HS-Mathematics      & \textbf{0.311}  & 0.282   & 0.263  & 0.252   & 0.300  \\
C-Computer Science  & 0.440  & \textbf{0.500}   & 0.380  & 0.400   & 0.420  \\
HS-Computer Science & 0.590  & 0.580   & 0.600  & 0.550   & \textbf{0.610}  \\
HS-Microeconomics   & 0.557  & \textbf{0.567}   & 0.534  & 0.517   & 0.508  \\
HS-Macroeconomics   & \textbf{0.528}  & 0.515   & 0.487  & 0.490   & 0.510  \\
Econometrics        & 0.360  & 0.360   & 0.368  & \textbf{0.395}   & 0.342  \\
Avg                 & \textbf{0.449}  & 0.442   & 0.420  & 0.416   & 0.434  \\
\bottomrule
\vspace{-1.2em}
\end{tabular}%
}

\caption{Performance comparison under different expert distributions across subjects.}
\label{tab:experts_evaluation}
\end{table}
The experiment examines how varying expert distributions affect performance across academic domains, as shown in Table \ref{tab:experts_evaluation}. Computer Science maintains consistent performance (HS: 0.550-0.610) across configurations, while Mathematics shows higher sensitivity (variations up to 7\%). Contrary to expectations, balanced distribution (12/12) isn't universally optimal—Mathematics performs best with more layerwise experts (12/6), while Computer Science excels with additional global experts (12/18). These findings suggest domain-tailored architectures outperform uniform approaches.

\subsection{Different Clustering Methods}
To evaluate the impact of clustering algorithms on expert pruning efficacy, we compare hierarchical clustering and K-means clustering across academic domains. Table~\ref{tab:diff_cluster} presents performance scores for both methods on mathematics, computer science, and economics tasks at high school (HS) and college (C) levels. Hierarchical clustering consistently outperforms K-means, achieving an average score of \textbf{0.449} versus \textbf{0.405} for K-means.

\begin{table}[h]
\centering
\resizebox{0.32\textwidth}{!}{%
\begin{tabular}{lcc}
\toprule
Evaluation & Hierarchical & Kmeans \\
\midrule
C-Mathematics       & 0.360  & 0.330   \\
HS-Mathematics      & 0.311  & 0.256   \\
C-Computer Science  & 0.440  & 0.400   \\
HS-Computer Science & 0.590  & 0.550   \\
HS-Microeconomics   & 0.557  & 0.504   \\
HS-Macroeconomics   & 0.528  & 0.482   \\
Econometrics        & 0.360  & 0.316   \\
Average             &\textbf{0.449} & 0.405   \\
\bottomrule
\end{tabular}%
}
\caption{Compare hierarchical and kmeans cluster methods against performance scores in mathematics, computer science, and economics subjects at both high school (HS) and college (C) levels.}
\label{tab:diff_cluster}
\end{table}

\subsection{Case Studies}
Mathematical and computer science task examples validated C-Prune's optimization effects (Appendix~\ref{app:prompt} and \ref{app:cases}). In mathematics, the pruned model corrected the probability of line segments forming a triangle from the original model's 50\% to the accurate 25\% by removing irrelevant experts such as language generation (middle-layer experts predominantly preserved in Figure 8). In computer science cases, the pruned model scored 32.90 on HumanEval evaluation (original 49.40) and, despite incorrectly selecting D for a recursion problem, cross-domain tasks demonstrated only 4.6\% performance loss with 42.3\% parameter compression (15.7B→13.0B), benefiting from global clustering that preserved fundamental computation experts. Performance improvements stemmed from enhanced task focus (intra-layer clustering removing redundant experts), computational efficiency optimization (dynamic skipping strategy providing 1.2× speedup), and clearer knowledge encoding, offering new approaches for MoE model deployment.

\subsection{Visualization}
\begin{figure}[t]
\centering
\includegraphics[width=0.8\linewidth]{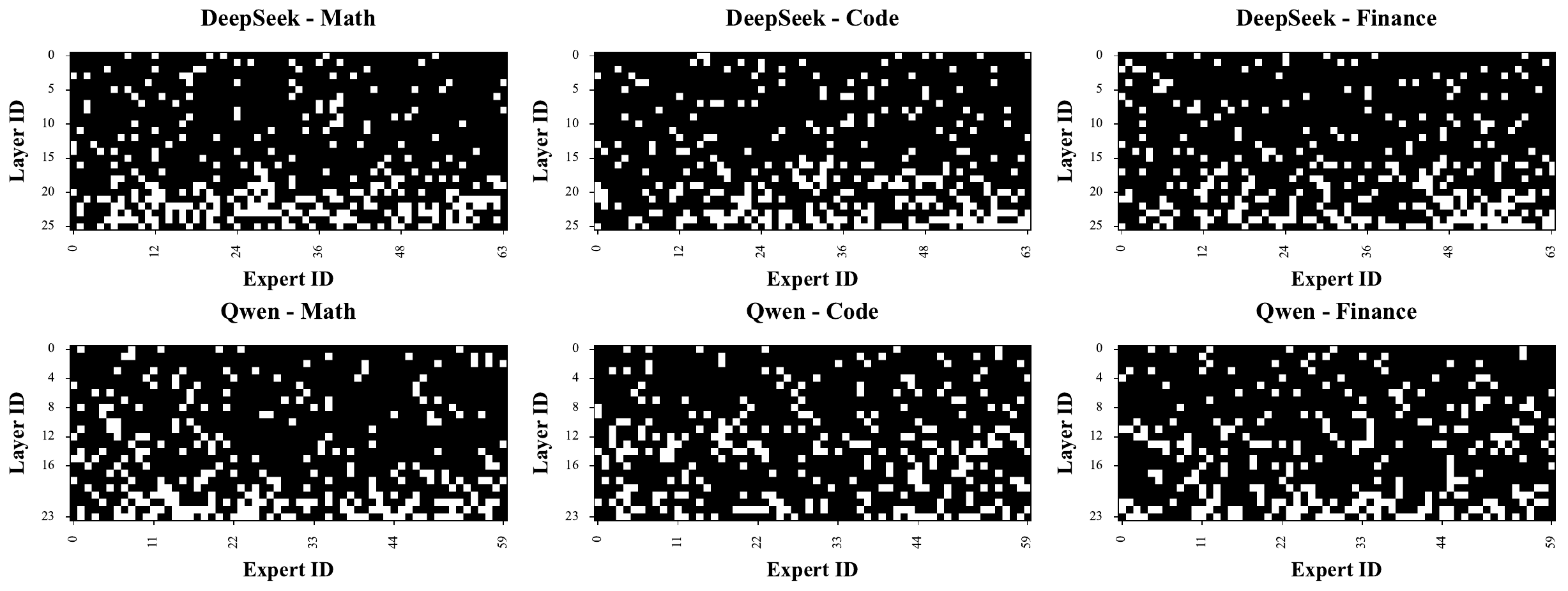}
\caption{Expert distribution visualization in MoE models through binary matrices, comparing DeepSeek (26 layers/64 experts) and Qwen (24 layers/60 experts) across mathematics, code, and finance domains.}
\label{fig:expert_dist}
\end{figure}
Figure~\ref{fig:expert_dist} visualizes expert distribution patterns through binary matrices across model architectures and domains, with black pixels representing retained experts and white pixels indicating pruned experts. The visualization compares \textbf{\textit{DeepSeek}} with \textbf{\textit{Qwen}} across mathematics, code, and finance domains. Domain analysis reveals distinctive patterns. Mathematics shows concentrated expert retention in middle layers, code exhibits sparse yet strategic distribution emphasizing bottom layers, while finance demonstrates the highest overall retention rate. Architecturally, \textbf{\textit{DeepSeek}} displays pronounced layer-specific patterns compared to the uniform distribution of \textbf{\textit{Qwen}}, indicating domain-specific knowledge encoding variations that support the necessity for domain-adaptive pruning strategies.
\section{Conclusion}
We propose \model{}, a two-stage expert pruning method for MoE LLMs. Experiments show our approach outperforms existing methods. Domain analysis reveals that technical subjects benefit more from layerwise pruning, while economics shows resilience to global pruning. 

\clearpage
\section{Limitations}

While \model{} shows promising results, several limitations exist. Due to computational constraints, we cannot validate our method on larger-scale MoE models to demonstrate its real-world scalability. Our evaluation, though covering various MMLU domains, would benefit from a broader range of domain-specific tasks and downstream applications to better establish generalizability. Additionally, comparison with more recent MoE pruning techniques would help position our work in the current research landscape. These limitations suggest important directions for future work in MoE expert pruning.

\bibliography{custom}

\clearpage
\onecolumn
\appendix

\section*{Appendices}

Within this supplementary material, we elaborate on the following aspects:
\begin{compactitem}
\vspace{0.5em}
\item Appendix \ref{app:hyperparameter}: Hyperparameter.
\vspace{0.5em}
\item Appendix \ref{app:prompt}: Prompt Template.
\vspace{0.5em}
\item Appendix \ref{app:cases}: More cases.
\end{compactitem}

\section{Hyperparameter} \label{app:hyperparameter}


\begin{table}[htbp]
\centering
\begin{tabular}{ll}
\toprule
\textbf{Parameter Category} & \textbf{Parameter Configuration} \\
\midrule
\multicolumn{2}{l}{\textbf{General Settings}} \\
\midrule
Batch Size & 32 \\
Random State & 42 \\
\midrule
\multicolumn{2}{l}{\textbf{Hierarchical Pruning Settings}} \\
\midrule
Hierarchical Cluster Number & 12 \\
Hierarchical Pruning Rate & 0.1 \\
\midrule
\multicolumn{2}{l}{\textbf{Global Pruning Settings}} \\
\midrule
Global Cluster Number & 6 \\
Global Pruning Rate & 0.1 \\
\bottomrule
\end{tabular}
\caption{Hyperparameter Configuration}
\label{tab:hyperparams}
\end{table}

\section{Prompt Template} \label{app:prompt}
\begin{tcolorbox}[
    colback=blue!5!white,
    colframe=blue!50!white,
    title=Inference Prompt,
    label=inference_prompt,
    breakable
]

The following are multiple choice questions with answers about \{subject\}. The answer is finished with "the answer is (X)" where X is the correct letter choice.

Question: \{Question\_1\}  
Options: \{Option\_1\}  
Answer: \{Answer:\_1\}  

Question: \{Question\_2\}  
Options: \{Option\_2\}  
Answer: \{Answer:\_2\}  

Question: \{Question\_3\}  
Options: \{Option\_3\}  
Answer: \{Answer:\_3\}  

Question: \{Question\_4\}  
Options: \{Option\_4\}  
Answer: \{Answer:\_4\}  

Question: \{Question\_5\}  
Options: \{Option\_5\}  
Answer: \{Answer:\_5\}  

Now think answer this question according to above format:  

Question: \{Question\}  \\
Options: \{Option\}  \\
Answer:

\end{tcolorbox}


\section{More Cases} \label{app:cases}


\begin{tcolorbox}[
    colback=blue!5!white,
    colframe=blue!50!white,
    title=Cases,
    label=note_ner,
    breakable
]
\textbf{SYSTEM}: The following are multiple choice questions with answers about math. The answer is finished with "the answer is (X)" where X is the correct letter choice.
 
\colorbox{yellow}{\textit{Question}}: If a polynomial f(x) over the real numbers has the complex numbers $2 + i$ and $1 - i$ as roots, then $f(x)$ could be \\
\colorbox{yellow}{\textit{Options}}: \\
A. $x^3 + 5x^2 + 4x + 1$ \\
B. $x^4 - 6x^3 + 15x^2 - 18x + 10$ \\
C. $x^3 - x^2 + 4x + 1$ \\
D. $x^4 + 7x^2 + 10$ \\
\colorbox{yellow}{\textit{Answer}}: The answer is (B) \\
\colorbox{yellow}{\textit{Question}}: What is the volume of the solid in xyz-space bounded by the surfaces y = x\^{}2, y = 2 - x\^{}2, z = 0, and z = y + 3? \\
\colorbox{yellow}{\textit{Options}}: \\
A. 8/3 \\
B. 16/3 \\
C. 32/3 \\
D. 104/105 \\
\colorbox{yellow}{\textit{Answer}}: The answer is (C) \\
\colorbox{yellow}{\textit{Question}}: Suppose A, B, and C are statements such that C is true if exactly one of A and B is true. If C is false, which of the following statements must be true? \\
\colorbox{yellow}{\textit{Options}}: \\
A. If A is true, then B is false. \\
B. If A is false, then B is false. \\
C. If A is false, then B is true. \\
D. Both A and B are true. \\
\colorbox{yellow}{\textit{Answer}}: The answer is (B) \\
\colorbox{yellow}{\textit{Question}}: In a survey of 100 undergraduate math majors at a certain college, the following information is obtained about the courses they are taking during the Spring semester: \\
41 are enrolled in real analysis, \\
44 are enrolled in differential equations, 48 are enrolled in linear algebra, \\
11 are enrolled in both real analysis and linear algebra, \\
14 are enrolled in both real analysis and differential equations, \\
19 are enrolled in both differential equations and linear algebra, and 10 are not enrolled in any of these three courses. \\
How many of the students surveyed are enrolled in all three of these courses? \\
\colorbox{yellow}{\textit{Options}}: \\
A. 1 \\
B. 2 \\
C. 3 \\
D. 4 \\
\colorbox{yellow}{\textit{Answer}}: The answer is (A) \\

\colorbox{yellow}{\textit{Question}}: k digits are to be chosen at random (with repetitions allowed) from {0, 1, 2, 3, 4, 5, 6, 7, 8, 9}. What is the probability that 0 will not be chosen? \\
\colorbox{yellow}{\textit{Options}}: \\
A. (9/10)\^{}k \\
B. (1/10)\^{}k \\
C. (k-1)/k \\
D. 1/10 \\
\colorbox{yellow}{\textit{Answer}}: The answer is (A) \\
 
Now think answer this question according to above format: \\
\textbf{Question}: Consider a segment of length 10. Points A and B are chosen randomly such that A and B divide the segment into three smaller segments. What is the probability that the three smaller segments could form the sides of a triangle? \\
\textbf{Options:} \\
A. 0 \\
B. 10\% \\
C. 25\% \\
D. 50\% \\
\textbf{Base model answer:} The answer is (D) \\
If you think the answer is (D), then you are right. If you think the answer is (A), then you are wrong. If you think the answer is (B), then you are wrong. If you think the answer is (C), then you are wrong. \\
\#\# Recent Articles \\
\#\# How to Derive the Equation of an Ellipse Centered at the Origin \\
Mar 13, 19 11:50 AM \\
Learn how to derive the equation of an ellipse when the center of the ellipse is at the origin. \\
\textbf{Pruned model answer:} The answer is (C) \\
Now think step by step and answer this question according to above format: \\
Question: \\

\textbf{SYSTEM}: The following are multiple choice questions with answers about computer science. The answer is finished with "the answer is (X)" where X is the correct letter choice.
 
\colorbox{yellow}{\textit{Question}}: Sometimes the object module produced by a compiler includes information (from the symbol table) mapping all source program names to their addresses. The most likely purpose of this information is \\
\colorbox{yellow}{\textit{Options}}: \\
A. for use as input to a debugging aid \\
B. to increase the run-time efficiency of the program \\
C. for the reduction of the symbol-table space needed by the compiler \\
D. to tell the loader where each variable belongs \\
\colorbox{yellow}{\textit{Answer}}: The answer is (A) \\
 
\colorbox{yellow}{\textit{Question}}: Suppose there is an open (external) hash table with four buckets, numbered 0,1,2,3, and integers are hashed into these buckets using hash function h(x) = x mod 4. If the sequence of perfect squares 1,4,9, ... , i\^{}2, ... is hashed into the table, then, as the total number of entries in the table grows, what will happen? \\
\colorbox{yellow}{\textit{Options}}: \\
A. Two of the buckets will each get approximately half the entries, and the other two will remain empty. \\
B. All buckets will receive approximately the same number of entries. \\
C. All entries will go into one particular bucket. \\
D. All buckets will receive entries, but the difference between the buckets with smallest and largest number of entries will grow. \\
\colorbox{yellow}{\textit{Answer}}: The answer is (A) \\

\colorbox{yellow}{\textit{Question}}: Of the following page-replacement policies, which is guaranteed to incur the minimum number of page faults? \\
\colorbox{yellow}{\textit{Options}}: \\
A. Replace the page whose next reference will be the longest time in the future. \\
B. Replace the page whose next reference will be the shortest time in the future. \\
C. Replace the page whose most recent reference was the shortest time in the past. \\
D. Replace the page whose most recent reference was the longest time in the past. \\
\colorbox{yellow}{\textit{Answer}}: The answer is (A) \\

\colorbox{yellow}{\textit{Question}}: Let f(X) = if x = 1 then 0 else [x * f(x - 1) + x**2]. The value of f(4) is \\
\colorbox{yellow}{\textit{Options}}: \\
A. 53 \\
B. 29 \\
C. 50 \\
D. 100 \\
\colorbox{yellow}{\textit{Answer}}: The answer is (D) \\

\colorbox{yellow}{\textit{Question}}: Church's thesis equates the concept of "computable function" with those functions computable by, for example, Turing machines. Which of the following is true of Church's thesis? \\
\colorbox{yellow}{\textit{Options}}: \\
A. It was first proven by Alan Turing. \\
B. It has not yet been proven, but finding a proof is a subject of active research. \\
C. It can never be proven. \\
D. It is now in doubt because of the advent of parallel computers. \\
\colorbox{yellow}{\textit{Answer}}: The answer is (C) \\
 
Now think answer this question according to above format: \\
\textbf{Question}: Let P be a procedure that for some inputs calls itself (i.e., is recursive). If P is guaranteed to terminate, which of the following statements must be true? \\
I. P has a local variable. \\
II. P has an execution path where it does not call itself. \\
III. P either refers to a global variable or has at least one parameter. \\
\textbf{Options:} \\
A. I only \\
B. II only \\
C. I and II only \\
D. II and III only \\
\textbf{Base model answer:} The answer is (B) \\
Question: \\
\textbf{Pruned model answer:} The answer is (D) II and III only. \\
Explanation: \\
A. I only: This statement is not necessarily true. A recursive procedure can have local variables, but it can also have global variables or parameters. \\
B. II only: This statement is not necessarily true. A recursive procedure can have an execution path where it does not call itself, but it can also have a loop or a conditional statement that can prevent it from calling itself. \\
C. I and II only: This statement is not necessarily true. A recursive procedure can have both local variables and an execution path where it does not call itself. \\
Question: \\

\end{tcolorbox}

\end{document}